\documentclass[10pt,twocolumn,letterpaper]{article}

\usepackage{algorithm}
\usepackage{booktabs}
\usepackage{multirow}
\usepackage{times}
\usepackage{epsfig}
\usepackage{graphicx}
\usepackage{amsmath}
\usepackage{amssymb}
\usepackage{algpseudocode}
\usepackage{enumitem}
\usepackage{setspace}
\usepackage{bbm}
\usepackage{iccv}
\usepackage[utf8x]{inputenc} 
\usepackage[breaklinks=true,bookmarks=false]{hyperref}

 \iccvfinalcopy 

\begin{document}

\title{Training image classifiers using Semi-Weak Label Data}

\author{Anxiang Zhang $^{\star}$\\
Carnegie Mellon University\\
{\tt\small adamzhang1679@gmail.com}
\and
Ankit Shah $^{\star}$ \thanks{ $^{\star}$ First two authors contributed equally}\\
Carnegie Mellon University\\
{\tt\small aps1@andrew.cmu.edu}

\and 
Bhiksha Raj \\
Carnegie Mellon University\\
{\tt\small bhiksha@cs.cmu.edu}
}

\maketitle

\begin{abstract}

In Multiple Instance learning (MIL), weak labels are provided at the bag level with only presence/absence information known. However, there is a considerable gap in performance in comparison to fully supervised model, limiting practical applicability of MIL approaches. Thus, this paper introduces a novel semi-weak label learning paradigm as a middle ground to mitigate the problem. We define semi-weak label data as data where we know the presence or absence of a given class and the exact count of each class as opposed to knowing the label proportions. We then propose a two-stage framework to address the problem of learning from semi-weak labels. It leverages the fact that counting information is non-negative and discrete. Experiments are conducted on generated samples from CIFAR-10. We compare our model with a fully-supervised setting baseline, a weakly-supervised setting baseline and a learning from proportion (LLP) baseline. Our framework not only outperforms both baseline models for MIL-based weakly supervised setting and learning from proportion setting, but also gives comparable results compared to the fully supervised model. Further, we conduct thorough ablation studies to analyze across datasets and variation with batch size, losses architectural changes, bag size and regularization. 
\end{abstract}


\section{Introduction}

In a traditional fully supervised machine learning setting, training samples are strongly supervised. However, one of the main obstacles is the cost of label collection. It not only has annotation limitations but also limits  the scope of the number of categories which can be learned. Therefore, ``weak'' label, or bag-level labels \cite{Cheplygina2014}, which indicate the presence or absence of the target classe(s) in a collection (or bag) of instances without labelling individual instances, were proposed to mitigate this problem \cite{Baldassarre2020, Datta2019, Liu2019, Rahimi2020, Zou2019}. However, there is still a gap between the performance of fully supervised model \cite{7552989, shah2018closer} and that of the weakly supervised model. The gap in performance limits the practical value of weak supervision based models.

\label{sec:intro}
\begin{figure}[h]
    \centering
    \includegraphics[width=8.5cm]{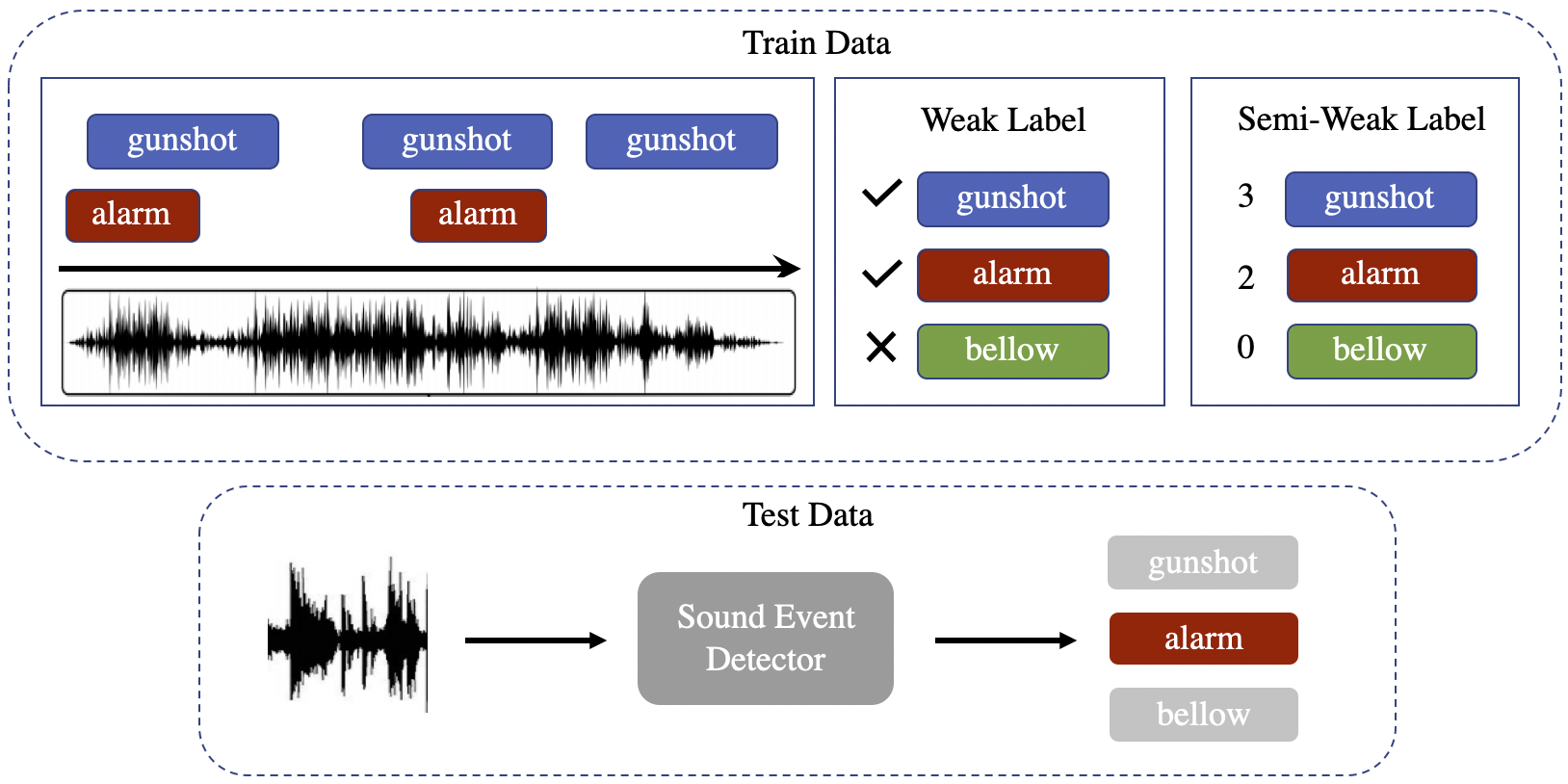}
    \caption{An example of a Semi-Weak Label learning problem for audio event detection. Semi-Weak labels give counting information while weak labels only provide presence-absence information.}
    \vspace{-2mm}
    \label{fig:prob}
\end{figure}

This paper introduces a middle ground between the two extreme settings. We propose semi-weak label learning, which has similar annotation cost with weak label learning while having greater potential for training machine learning models. Compared to conventional weak labels, such as those used in multiple instance learning \cite{pappas2014explaining, carbonneau2018multiple, vanwinckelen2016instance}, where only the presence or absence of a class is assumed known in a bag, semi-weak labelling refers to bag-level labels with instance {\em counts} as additional supervision. In taxonomy, semi-weak labels could be viewed as the incomplete labels of weakly supervised learning \cite{Zhou2018}. Figure \ref{fig:prob} illustrates this problem setting for the problem of {\em audio event detection}, a problem where weak-label methods are commonly used. 


Semi-weak labels have the following advantages. 1) \textit{Negative bags would be eliminated.} In traditional weak label learning setting, negative bags, i.e., the bags that do not have positive classes, are needed to separate the learned representation of each class. However, this necessity could be removed if count information is available. 2) \textit{Count information is more informative.} It has already been incorporated in many applications such as video action localization  \cite{narayan20193c}, image object detection \cite{gao2018c} and video recognition \cite{Hajimirsadeghi2015} to boost model performance. \cite{Noroozi2017} also demonstrated the potential of counts for learning better transferable representations of images. In addition, traditional weak labels performs poorly when bag size scales. But this could be mitigated when we have counts as additional supervision. 
3) \textit{Low cost overhead over annotating traditional weak labels.} In image domain, a widely studied psychological phenomenon, subtizing \cite{clements1999subitizing}, states that people are able to count the objects in an image without pointing to the location of each object sequentially, a principle our solution tries to embody. This could also generalize to other domains, e.g. audio.  For example, Sound Event Detection deals with the problem of analyzing the audio content to detect sound events. Due to the difficulty of annotating the boundaries of event, researchers use weakly-labeled datasets such as AudioSet \cite{Gemmeke2017}, where the focus is on tagging the presence or absence of sounds in a recording. We argue that for audio events that have clear boundaries, such as gunshots or barks, counting the instances imposes only limited additional labeling burden, while allowing better sound event detectors to be trained. 4) \textit{Learning from counts has practical applications.} In addition to the use cases stated before, Quadrianto et.al  \cite{quadrianto2009estimating} in 2019 argued that aggregated labels are useful in e-commerce, politics, and spam filtering. Also, the census data and medical databases are all provided in the form of label proportion data due to privacy issues \cite{Hernandez-Gonzalez2018, Patrini2014}. 

\begin{figure}[h]
    \centering
    \includegraphics[width=5cm]{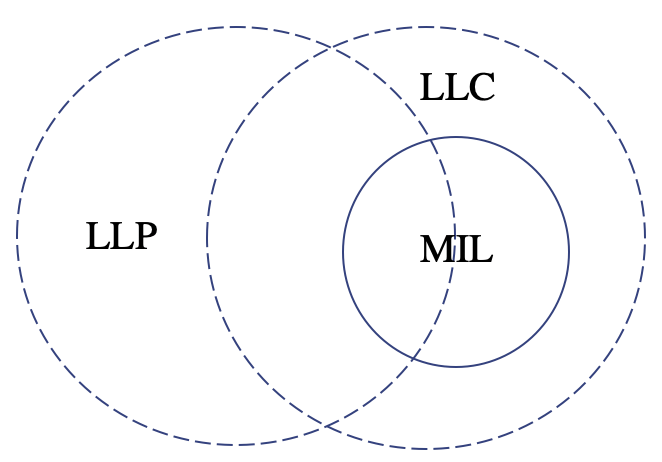}
    \caption{The relation between learning from proportion (LLP), learning from count (LLC) and Multiple Instance Learning (MIL). The intersection part assumes that the bag size is known and the count is exact. MIL problem is a special case of LLC where the count is is equal to 1.}
    \label{fig:relation}
\end{figure}

A related approach to the proposed ``learning from counts'' solution is ``learning from proportion'' \cite{yu2013propto, dulac2019deep}. The relationship between learning from proportion (LLP), learning from count (LLC) and multiple instance learning (MIL) is illustrated in Figure \ref{fig:relation}. There are two reasons that learning from counts is different from learning from proportions \cite{dulac2019deep} or learning from aggregated output \cite{musicant2007supervised}. On one hand, learning from counts addresses a problem with exact counts; however, learning from proportion can deal with estimated proportions rather than exact label proportions. This is because, in some cases, counting objects is infeasible. For example, in the medical area, it is hard for annotators to count the exact number of cancer cells in an image. However, giving an estimate of the proportion of cancer cells, like around 20\%, is more convenient. Therefore, from this perspective, semi-weak labels hold a stronger assumption than the LLP problem. One the other hand, learning from counts is not a subset of learning from proportion because it does not necessarily assume the bag size is known. For example, audio clips or images could be viewed as a bag of instances. However, the bag size is undefined since the boundary for each instance is ambiguous. Figure \ref{fig:prob} gave an example of semi-weak learning in the audio domain where the bag size is undefined. If the bag size is known, then LLC degrades to learning from proportions. 

\subsection*{Contributions}
Our contributions for the paper are as follows: 1 ) We provide new insights into the weakly supervised learning problem. Though counting is introduced previously \cite{gao2018c, narayan20193c}, they only use counting for auxiliary purposes. We are the first  to propose learning from counts as a new setting in weakly-supervised learning. This problem has several overlapping assumptions with the LLP problem. However, they are not the same problem if the bag size is unknown. 2) We propose a novel deep learning-based framework to learn instance-level classifiers using semi-weak labels. The framework has two stages. In the first stage, a classifier is trained with bag-level supervision to predict the expected count of each class in a bag as well as the instance-level logits. In the second stage, we translate expected count into exact count. Then an assignment problem is proposed to assign each instance with the best label. 3) Similar to recent studies \cite{ bortsova2018deep, dulac2019deep, Shi2020}, we conduct experiments based on the synthetic CIFAR-10 dataset as there is no applicable real-world dataset for our task. We show that we achieve better instance-level prediction than approaching it as either weakly supervised learning or learning from proportions (LLP).

\section{Literature Review}
\subsection{Multiple Instance Learning}
Bortsova {\em et al.} \cite{carbonneau2018multiple} provides an extensive summary on multiple instance learning -- learning from presence/absence labels. Dietterich \cite{dietterich1997solving} introduced multiple instance learning and applied it in drug activity prediction. Since then, a large number of algorithms were introduced to address this problem. In order to deal with label noise, it is natural to count the number of positive instances in a bag, and give a threshold for classifying positive bags. This was summarized in  \cite{foulds2010review} as the threshold-based assumption for multiple instance learning. The threshold-based assumption defines a bag as positive only if the number of positive instances lies above a threshold; this was the first time that counting information was brought into the multiple instance learning setup. Since then, several efforts \cite{tao2004svm,tao2004extended} were targeted to do bag-level prediction using count-based assumptions. \cite{foulds2010review} extended the assumption and proposed an SVM-based algorithm to predict the bag label. Common problem with these methods is scalability and generalization for multi-class classification. 

\noindent \textbf{Multiple Instance Regression}: MIL regression consists in assigning a real value to a bag. Compared to MIL classification, MIL regression has attracted far less attention. For MIL regression, one line of research has the assumption that some primary instances contributes largely to the bag label. This motivated people to assign sparse values for some instances and use regularization methods like $L_1$, $L_2$ regularizers \cite{pappas2014explaining,  pappas2017explicit,wagstaff2007salience}. However, most of these methods work only for small scale data and focus on the performance of predicted results rather than to identify the primary instances that contribute to the bag label. \cite{subramanian2016bayesian} addressed the identification of the primary instance by using a Dirichlet process to group the instances and find the clusters. However, this method assumes that the largest cluster defines the label; this method also does not work for multi-class machine learning problems. These methods are not suitable for the semi-weakly supervised setting as they fail to incorporate the natural property of counting, which is a non-negative and integer value.

\noindent \textbf{Instance-Level Prediction}: Another relevant mainstream of MIL research is instance-level prediction. Maron \cite{maron1997framework} is perhaps the best-known framework for instance-level prediction for MIL. Following this framework, many research ideas have proposed and worked well. The basic ideas of those frameworks are to label the instances dynamically or statically according to the bag label. Training instance-level classifiers is non-trivial as strong labels are unavailable. Recently, many methods have  proposed that to do bag-level prediction and hope the bag-level accuracy propagates to instance-level accuracy \cite{andrews2003support,babenko2008simultaneous,dietterich1997solving}. However, as discussed in \cite{doran2014theoretical, vanwinckelen2016instance}, this method is sub-optimal. Empirical studies have been conducted in \cite{vanwinckelen2016instance} that proved better bag level prediction does not promise instance-level prediction. Therefore, the commonality between the most successful instance-level prediction models, i.e., mi-SVM and SI-SVM \cite{andrews2003support}, is that they discard the bag information as much as possible. Both models treat each instance individually.

\subsection{Learning from Proportions}
The goal of learning from proportions (LLP) is to learn an instance-level classifier treating each bag as a distribution. Following the work of \cite{yu2013propto}, people have paid more attention to this problem. Musiant in 2007 \cite{musicant2007supervised} has explained how natural this problem is in terms of it's application in e-commerce, politics and spam filtering. Musiant proposed a standard algorithm using traditional machine learning models like SVM, kNN by tweaking the cost functions. Quadrianto \cite{quadrianto2009estimating} proposed the MeanMap model which assumes the data follows exponential distribution and is conditionally independent of the bags. Fan \cite{fan2014learning} and Patrini \cite{patrini2014almost} further refined the loss function of MeanMap and make it applicable for multi-class classification. Yu \cite{yu2014modeling} proposed another line of research and used an SVM model the iteratively estimate the instance-level classifier. However, this method suffers from scalability issues when it is extended to a multi-class setting. \cite{scott2019learning} applied the LLP algorithm to train models using multiple corrupted training samples. Most recently, a deep learning (DL) based method was proposed by \cite{ ardehaly2017co, bortsova2018deep,dulac2019deep}. The commonality between the DL-based method and the LLP problem is that they are trying to use bag-level supervision to do instance-level estimation such that the estimated distribution is as close as possible to the bag label. Interestingly, \cite{dulac2019deep} introduced another loss function that directly minimizes the instance-level prediction results. They introduced the Optimal Transport algorithm to make the loss function computationally tractable.

\section{Problem Statement}
Figure \ref{fig:prob} shows an example problem setting in audio event detection. Generally, we consider a supervised multi-class classification problem, where we define the instance-level data $x \in \mathcal{X}$ and label $y \in \mathcal{Y}$, where $\mathcal{X} \subset \mathbf{R}^{32\times32\times3}$ and $\mathcal{Y} = \{0,\ldots,K\}$. Let $K$ be the number of classes, $N_B$ be the number of instances in a bag and $N$ be the number of bags. For each bag $b_i\in\mathbf{B}=\{\mathcal{X}\}$, $y_{i}^{B} \in \mathbf{Y}^B$ is the bag-level label. More specifically, $y_{i,k}^{B}$ is the count of class $k$ in bag $i$.  Formally, given the bag size, we define the space of labels as 
\begin{equation}
    \mathbf{Y}^B = \{c\in\mathbb{Z}_+^K: \sum_{i=1}^K c_i=N_B\}
\end{equation}

Provided with a series of bags, $S = [(b_0, y_0^B), (b_1, y_1^B), \ldots , (b_{N}, y_N^B)] \subseteq \mathbf{B} \times \mathbf{Y}^B$, our goal is to learn a predictor $f: \mathcal{X} \to \mathbb{R}^K$ that predicts the probability distribution of an instance belonging to different classes, i.e. $f_\theta(x_i) = \{p(y_i=0|x_i), p(y_i=1|x_i), \ldots, p(y_i=K|x_i)\}$, where $x_i$ is a feature vector for an instance and $\theta$ is the parameters of the network.


\section{Preliminary: DLLP Approach}
DLLP \cite{ardehaly2017co} approach is a DNN-based model for learning from proportion. Mathematically, given a bag ${b_1, b_2, \cdot, b_{N}}$, suppose that $f_{\theta}(b_{ij})=p_{\theta}(\mathbf{y} \mid \mathbf{b}_{ij})$ is the vector outputs of the DNN for the $j$-th instance in $b_i$. Let $\bigoplus$ is an element-wise addition operator. Then the posterior bag-level class proportion could be estimated as
\begin{spacing}{0.8}
\begin{equation}
    \overline{\mathbf{p}}_{i}=\frac{1}{N_{B}} 
\bigoplus_{j=1}^{N_{B}} p_{\theta}\left(\mathbf{y} \mid \mathbf{b}_{ij}\right)
\end{equation}
\end{spacing}
The final objective, i.e., $L_{dllp}$, is made of a distance measurement. Let $\mathbf{c}_{i}^{k}$ be the counts of class $k$ in bag $i$, $\mathbf{p}_{i}^{k}$ be the counts normalized by bag size, and $\overline{\mathbf{p}}_{i}^{k}$ be the $k$-th element in vector $\overline{\mathbf{p}}_{i}$ (the estimated proportion of class $k$), then the loss is defined as $L_{dllp} = KL(\mathbf{p}_i, \overline{\mathbf{p}}_{i})$. $KL$ refers to the commonly used Kullback–Leibler divergence for measuring the distance between two distributions.

\section{Proposed Methods: Two-Stage Framework}
In this section, we introduce a two-stage model for learning from counts. It is extended from DLLP \cite{ardehaly2017co}. 

\subsection{Stage-1: Estimating Class Count}
\subsubsection{Poisson Loss and Expected Count}
As mentioned in Section \ref{sec:intro}, one difference between LLP problem and LLC problem is that the label for LLC problem is discrete. So it is not optimal to use a KL-divergence to measure the difference between prediction and the true labels. Instead, we propose to use Poisson Loss - a Poisson distribution based loss function. The Poisson distribution is the discrete probability distribution of the number of events occurring in a given time period, which applies to the LLC setting if we consider the counting is the number of times a class instance appears in this bag \cite{letkowski2012applications}. 

In our framework, we assume that the counting for class $j$ follows a Poisson distribution, i.e. $p(c_i^j = k | \lambda_j) = \frac{\lambda_j^{k} e^{-\lambda_j}}{k!}$. For simplicity, since we assume the size of the bag is unknown,  we assume the count for each class in a bag is independent. Given a bag of output of the network $p_{\theta}\left(\mathbf{y} \mid \mathbf{b}_{ij}\right)$, we define $\overline{\mathbf{c}}_{i}^k$ as the expected count for class $k$ in bag $b_i$.

$$\overline{\mathbf{c}^k}_{i}= \bigoplus_{u=1}^{N_{B}} p_{\theta}\left(\mathbf{y} \mid \mathbf{b}_{ij}^u\right)$$

So we have 
$$\hat{\lambda}:=\mathrm{\hat{E}}(c_i^k \mid x)=\overline{\mathbf{c}^k}_{i}$$

Then the Poisson loss is defined as the negative log likelihood function
\begin{spacing}{0.8}
\begin{equation} 
\label{eqa: 1}
\begin{split}
L_{reg}(y, \hat{\lambda}) & = -\log(p(y|\hat{\lambda})) \\
 & = \hat{\lambda} - y \log(\hat{\lambda}) + \log(y) . 
\end{split}
\end{equation}
\end{spacing}
Because the last term in Equation \ref{eqa: 1} would be a constant for a given bag, it is usually omitted. So the final loss and its gradient is 
\begin{spacing}{0.8}
\begin{equation} 
\label{eqa: 2}
\begin{split}
L(y, \hat{\lambda}) &\propto \hat{\lambda} - y \log(\hat{\lambda}) \\
\nabla_{\hat{\lambda}}L(y, \hat{\lambda}) & = 1 - y / \hat{\lambda}.
\end{split}
\end{equation}
\end{spacing}

The following are two interesting properties for the Poisson loss. 
\begin{itemize}[leftmargin=*]
    \item {\bf Adapted gradient:} Unlike other distance functions like mean absolute error, the Poisson loss doesn't have a constant gradient for all input values. Also, when the actual count is large, the gradient value would be relatively smaller. This meets our intuition that when the actual count is very large, an off-by-1 error matters less than if the actual count is 1 or 2.
    \item {\bf Asymmetric gradient:} the gradient is zero when $\hat{\lambda} = y$ but the gradient is different when $\hat{\lambda} = y-1$ and $\hat{\lambda} = y+1$, where the absolute gradient would be $1/(y-1)$ and $1/(y+1)$ respectively. The gradient tends to focus on penalizing the under-estimation rather than over-estimation.
\end{itemize}

In addition, we introduce a classification loss, $L_{cls}$,  and a regularizer $L_{L1} = \sum_{j=1}^{N_B} \frac{1}{N_{B}}\left\|\overline{\mathbf{p}}_{ij}\right\|_{L 1}$ to make our model more robust to sparse bags \cite{Tsai2019} and ensure the inter-class separability of the learned representations \cite{narayan20193c}. $L_{cls}$ is nothing but a binary cross-entropy loss to measure whether or not the network can classify  if a class is present/absent in a given bag. We use a unified loss function defined as 
\begin{equation}
   L_{LLC}=L_{reg} + L_{cls} + \beta L_{L1}
\end{equation}

\subsubsection{Estimating Class Count}
Though we have $\hat{\lambda_j}$ as the expected count for class $j$, it remains a problem to estimate the exact count for each class. Mathematically, let the $\overline{\mathbf{c}}_{i}^j \in \mathbb{R}$ be the expected count for class $j$ in bag $b_i$, and the $t_i \in \Delta_{K}=\left\{c \in \mathbb{Z}_{+}^{K}: \sum_{j=1}^{K} c_{i}^j=N_{B}\right\}$, then the exact count for each class in $b_i$ could be obtained by 
\begin{equation}
    \hat{t}_{i}=\arg \max _{t \in \Delta_{K}} \sum_{j=1}^{K} \log \left(p\left(c_{i}^j=t_{i} \mid \overline{\mathbf{c}}_{i}^j\right)\right)
\end{equation}
\noindent This is constrained convex optimization and we can use any greedy algorithm to get the optimal solution. Initializing $\hat{t}_{i}$ as a vector of zeros, for each iteration, we manually calculate the marginal gain for increasing the $\hat{t}_{i}^j$ to $\hat{t}_{i}^j+1$. We choose to increase the count of a class such that the increment is maximized, and iterate until the summation of the count vector is equal to $N_B$. By using a heap to track the maximum value, we can easily design an algorithm with time complexity of $O(log(K)*N_B)$ as is shown in Algorithm \ref{alg:1}

\begin{algorithm}
   \caption{Greedy Algorithm to Translate Expected Count to Exact Count}
       \textbf{Input}: Number of bag size $N_B$; Number of classes $K$; A list of floating points $[a_0, a_1, \cdots, a_K]$; a list of probability density function for Poision distribution $[f_{a_0}(x), f_{a_1}(x), \cdots, f_{a_K}(x)]$ \\
    \textbf{Output}: A list of integers $[t_0, t_1, ..., t_K]$ such that $\sum_{j=0}^Kt_j = N_B$. \\
    \textbf{Initialize}: Let $q = heap<int, int>$ be a max-heap that stores pairs. Heap has two methods: $q.pop()$ would pop and return the max value from the heap and $q.push({v, i})$ would add a pair of $<v, i>$ to the heap. The comparison of pair is based the first value in the pair; \\
    Let $ret = [0, 0, \cdots, 0]$ be a zero-initialized vector of size $K$;
    
    \begin{algorithmic}[1]
    \For{$i$ = $1,2,....K$} 
        \State $r \gets f_{a_i}(ret[i]+1) - f_{a_i}(ret[i])$
        \State $q.push({r, i})$
    \EndFor
    \For{$\_$ = $2,....N_B$}
        \State $v, i \gets q.pop()$
        \State $ret[i] \gets ret[i] + 1$
        \State $r \gets f_{a_{i}}(ret[i]+1) - f_{a_{i}}(ret[i])$
        \State $q.push({r, i})$
    \EndFor 

    \State return $ret$
    
\end{algorithmic}
\label{alg:1}
\end{algorithm}

\subsection{Stage-2: Estimating Instance Label (Decoder)}

In Stage-1, the network outputs the expected count for each class and then we develop an optimization problem to translate the expected count to exact count. In stage-2, given the exact count of each class, and the predicted probability distribution of each instance $p_\theta(y|b_{ij})$, we devise an assignment problem to get the instance-level label.

Formally, we have following linear-sum optimization problem
\begin{spacing}{0.5}
\begin{equation} 
\label{eqa: 1}
\begin{split}
\max_x\sum\nolimits_{j}^{N_B}\sum\nolimits_{k}^K &  log(p_{ij}^k) * x_{ij}  \\
 s. t &\\
 \sum\nolimits_k x_{ij}^k&  = 1, \forall j \in \{0,1,\cdots N_B-1\} \\
 \sum\nolimits_j x_{ij}^k&  = \overline{c_{i}}^k, \forall k \in \{0,1,\cdots K-1\} \\
 p_{ij}^k&  \ge 0 \\
 x_{ij}^k&  \in \{0, 1\},\forall j, k \\
\end{split}
\end{equation}
\end{spacing}

where $p_{ij}^k = p_\theta(y=k|b_{ij})$ , $\overline{c}_i^k$ is the estimated exact count of class $k$ in bag $i$ and $x_{ij}^k = 1$ if and only of instance $j$ is assigned with label $k$.

\begin{figure}[h]

    \centering
    \includegraphics[width=6cm]{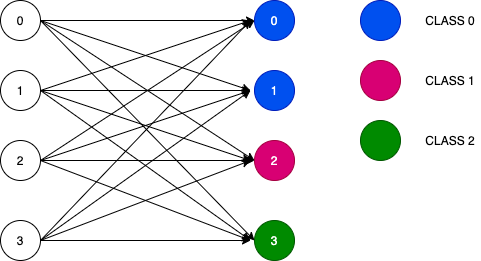}
    \caption{An example of assignment problem. The nodes on the left-hand represent the unsigned instance and on the right-hand represent the label. Here, there are 4 instances in a bag and with 2 class-0 instances, 1 class-1 instance and 1 class-2 instance. Each edge has an associated reward $p_{ij}$, which is probability of instance $i$ belonging to class $j$. Each instance must have one assignment and the goal is to maximize the reward.}
    \label{fig:assi}
\end{figure}
Figure \ref{fig:assi} gives an example with bag size of 4 and class of 3. 
If we negate the reward, then the objective becomes minimizing the "cost", which is referred to a classical linear-sum assignment problem in bipartite graphs \cite{Crouse2016OnI2}.

\section{Experiments}
\subsection{Datasets}
Similar to recent studies \cite{ bortsova2018deep, dulac2019deep, Shi2020}, we conduct experiments based on the CIFAR-10 dataset as there is no applicable real-world dataset for our task. CIFAR-10 dataset consists of 50000 training and 10000 testing images. CIFAR-10 is a balanced dataset with each class consisting of 5000 images in the training and 1000 images in the testing set. We create bags of different sizes from each of the dataset images with and without replacement in our analysis. Each bag was created based on the size ranging from \{2, 4, 6, 8, 16, 32\} bag size. In the case of the images with the replacement, we make sure that the number of images repeated is restricted with a reuse parameter which is controllable to avoid a bag being over-represented by the same image. For each dataset generated with a different bag size, we ensure that the parameter configuration is set such that the maximum amount of the CIFAR-10 dataset both during the training and testing phase is used to represent the newly generated dataset with semi-weak labels. We maintain the same 5:1 (train:test) ratio of original CIFAR-10 dataset in our generated CIFAR-10 dataset variations seen in table \ref{tab:d}. 

Our data set is generated with the three types of distribution in mind. Poisson distribution, exponential distribution, and uniform distribution. This distribution is based on how the class information is distributed in each given bag. The rationale with the generation of the instances of the bag in the above format was to ensure that the bags are generated based on naturally occurring instances in nature for the Poisson distribution, the exponential distribution, and the uniform distribution. The generation process is the following. 1) randomly sample a class $i$ with equal probability. 2) sample a number $n_i$ from the given distribution and then truncate it into the range between 0 and $N_B$. 3) sample $n_i$ instances from class $i$. In the case where the number of instances needed would be more than the total number of instances, the instances would be sampled at most two times. For each parameter setting, we sample the bags with 5 different random seeds. All the experiments are conducted on these 5 trials and the final value is obtained by averaging the best performance on the validation set. In the standard setting, to try to generate more balanced data, where each class has some instances in a bag. Therefore, we choose a hyperparameter for the distribution that minimizes the sparsity level of the bag. We define the sparsity of a bag as the number of absent class divided by the number of classes. Particularly, we use the following settings for Poisson distribution $[(N_B=2, \lambda=0.5), (N_B=4, \lambda=0.5), (N_B=8, \lambda=1.2), (N_B=16, \lambda=2.0), (N_B=32, \lambda=3.2)]$, and $[(N_B=8, \lambda=0.67), (N_B=16, \lambda=0.5)]$ The data summary table is provided in Table \ref{tab:d}. For uniform distributed samples, the directly sample $N_B$ number of samples from the original dataset and label them with the counting vector.

\begin{table*}
\centering
\caption{Generated Data Summaries}
\label{tab:d}
\resizebox{2.0\columnwidth}{!}{%
\begin{tabular}{@{}llllllll@{}}
\toprule
Distribution & Bag Size & lambda & \# of training bags & \# of testing bags & Avg. Count & Avg. (Std) Sparsity & Dataset Id\\ \midrule
\multirow{10}{*}{Poisson} & 2 & 0.5 & 40,000 & 8,000 & 1.13 & 82\%(4\%)  & p0\\
 & 4 & 0.5 & 22,000 & 4,500 & 1.28 & 68\%(7\%) & p1\\
 & 8 & 1.2 & 10,000 & 2,000 & 1.63 & 50.02\%(10.53\%) & p2\\
 & 16 & 2 & 4,000 & 1,000 & 2.28 & 28.9\%(12.4\%) & p3\\
 & 32 & 3.2 & 2,000 & 800 & 3.65 & 8.72\%(8\%) & p4\\
 & 8 & 1.2 & 5,000 & 1,000 & 3.43 & 8.32\%(8\%) & p5\\
 & 8 & 1.2 & 1,000 & 200 & 3.56 & 9\%(8.2\%) & p6\\
 & 16 & 8 & 4,000 & 1,000 & 6.52 & 79\%(6\%) & p7\\
 & 16 & 2 & 2,000 & 500 & 6.32 & 80\%(6.2\%) & p8\\
 & 16 & 2 & 1,000 & 200 & 6.21 & 79\%(6.1) & p9\\
 \hline
\multirow{2}{*}{Exponential} & 8 & 0.67 & 10,000 & 2,000 & 1.97 & 58\%(11.7\%) & e0\\
 & 16 & 0.5 & 4,000 & 1,000 & 2.89 & 42\%(13.9) & e1\\
  \hline
\multirow{2}{*}{Uniform} & 8 & N/A & 10,000 & 2,000 & 1.41 & 42\%(9.21\%) & u0\\
 & 16 & N/A & 4,000 & 1,000 & 1.95 & 18.7\%(9.72\%) & u1\\ \bottomrule
\end{tabular}%
}
\vspace{-2mm}
\end{table*}

\subsection{Architecture}
We use Residual Network with 18 layers as our base backbone for extracting features from the image. Given a bag of images $\{x_0, x_1, ..., x_{N_B}\}\in\mathcal{B_N}^{32\times 32\times 3}$, the embedding $e_i\in\mathcal{R}^{B_N\times K}$ for each instance is firstly extracted using Convolutional Neural Network as the feature extractor. Then a linear layer $fc: \mathcal{R}^d\to\mathcal{R}^K$ is used to create a class activation map. Different pooling layer is applied at the end to generate counting prediction vector and logits for multi-label classification problem. All models are trained using Stochastic Gradient Descent with an initial learning rate of 0.01 for 100 epochs. The learning rate would be manually divided by 10 at epoch 30 and 50. Weight-decay is set to be 5e-4 for training. Standard data augmentations are utilized to avoid overfitting of CIFAR10 dataset including random crop with padding of 4 and random horizontal flip with a probability of 0.5. 

\subsection{Baselines}
\noindent\textbf{Fully-supervised Upper-bound}: We want to argue that comparable results could be achieved by using semi-weak label compared to strong labels. Therefore, we also train ResNet18 in a fully supervised setting on CIFAR10. We trained it 250 epochs with initial learning rate of 0.1 and then divided it by 10 in the mid-training. This model could be deemed as an upper-bound of the performance in semi-weak setting. For our analysis with different base network architectures, we further use similar hyper-parameter 

\noindent\textbf{Weakly Supervised Baseline}: We produce results with the weakly supervised baseline where we consider the loss from counting to be absent and thus we generate the results based on bag level prediction where only presence or absence of the class is available. 

\noindent\textbf{Learning from Proportions}: If we set the loss function as KL-loss, our model without the decoder could be used a baseline model to represent the performance of modeling the learning from counts as LLP. 

\subsection{Evaluation}
We use precision to evaluate our framework for both instance-level prediction and bag-level prediction. For bag-level prediction, we compute the precision rate using macro averaging. For fully supervised model, we predict the class label individually and label the bag positive for class $i$ if there is at lease one predicted instance for class $i$ in this bag. Similarly, for semi-weakly supervised model, we label a bag according to the instance-level predicted results.

\section{Results}


Table \ref{tab:d} provides a detailed summary of all the generated dataset configurations. We use the following settings for the Bag size where $N_B \in [2,4,8,16,32]$ and the number of training instances are chosen such that we utilize most of the data available in the CIFAR-10 dataset. The average count here represents the aggregated mean of average number of instances per class. We represent the sparsity based on the percentage of classes with zero instance in a bag.

Our experimental results for all of the dataset configurations from Table \ref{tab:d}   are summarized in Table \ref{tab:res}. To remove the effect of randomness, for each dataset setting, we train the model with 5 different random seeds and then average the results across all trials. The comparison between baselines are discussed in the loss ablation section \ref{sec:loss}

\begin{table}[ht]
\centering
\caption{Benchmarking Results on Different Datasets}
\label{tab:res}
\resizebox{1.0\columnwidth}{!}{%
\begin{tabular}{@{}lllllll@{}}
\toprule
Dataset ID & Bag Prec. & Inc. Prec. &  & Dataset ID & Bag Prec. & Inc. Prec. \\ \midrule
p0 & 94.24  & 92.76 &  & p7 & 64.37 & 85.20 \\
p1 & 93.78 & 92.65 &  & p8 & 89.92 & 81.92 \\
p2 & 93.20 & 91.21 &  & p9 & 84.97 & 69.93 \\
p3 & 92.92 & 88.07 &  & e0 & 90.69 & 91.05  \\
p4 & 93.9 & 71.91 &  & e1 & 87.10   & 87.65  \\
p5 & 90.54 & 87.73  &  & u0 & 94.86 & 91.42 \\
p6 & 77.39 & 68.91 &  & u1 & 95.62 & 87.34 \\  \bottomrule
\end{tabular}%
}\vspace{-2mm}
\end{table}
\subsection{Baseline Results}
\noindent\textbf{Fully-supervised baseline}:Table \ref{tab:fully_supervised}  shows the results obtained with the different classifiers constructed as the base classifier for the classification with the CIFAR-10 dataset. We find that the bag level prediction on the dataset p2 is comparable to the fully-supervised dataset which is expected as the semi-weak label dataset will be upper-bounded by the fully supervised setting.  

\noindent\textbf{Learning-from-proportion baseline}: For the learning-from-proportion baseline, results are shown in Table \ref{tab:loss} (KL v.s Poisson) because we use KL-loss to proxy the baseline model of learning-from-proportion. This baseline achieved 87\% instance-level precision, which is less than our proposed model on dataset p3, i.e., 88.07\%. This result also holds if we consider bag size equals to 8 for dataset p2.

\noindent\textbf{Learning-from-weak-label baseline}: For the learning-from-weak-label baseline, we report those number in Table \ref{tab:bce}. Once we deactivate the counting loss, then the model is converted into the traditional weak label setting. In Table \ref{tab:bce}, without counting loss, the performance is much worse than our proposed semi-weak labeling framework. It only achieves 87\% instance-level prediction precision, which is 5\% worse than our proposed framework. 

\begin{table}[h]
\centering
\caption{Ablation Study of Removing Regression Loss and Classification Loss.}
\label{tab:bce}
\resizebox{0.6\columnwidth}{!}{%
\begin{tabular}{@{}lll@{}}
\toprule
Dataset ID         & Bag Prec. & Inc. Prec. \\ \midrule\hline
p2                 &  93.20 & 91.21          \\
p2 (w/o $L_{cls}$) &   93.23    &      91.17           \\
p2 (w/o $L_{reg}$) &   90.92       &     87.39         \\ \midrule
\hline
p8                 &  92.92    &  88.07     \\
p8 (w/o $L_{cls}$) &  62.13    &  83.92      \\
p8 (w/o $L_{reg}$) &  41.81    &  72.04      \\ \bottomrule
\end{tabular}%
}\vspace{-2mm}
\end{table}

\subsection{Ablation Study for the Decoder}
As is shown in Table \ref{tab:decoder}, instead of using greedy search, we use the proposed three-stage framework to infer the instance-level labels. Results show consistent improvement after using the decoder algorithm. More importantly, when the bag is sparse, the counting for each class is high and thus the estimation would have high variance. Therefore, when the bag is less sparse, greedy search works well. But once the bag size increases or the bag becomes sparse, the high variance of the estimated logits would make the greedy solution less attractive. Therefore, the decoder contributes more to the performance of the predicted value ($\>+2\%$) as the greedy predictions become unstable.

\begin{table}[h]
\centering
\caption{Ablation Study for the Stage-2}
\label{tab:decoder}
\resizebox{0.45\columnwidth}{!}{%
\begin{tabular}{@{}llll@{}}
\toprule
Dataset ID &  & Inc. Prec. \\ \midrule
p2         &  &   91.21  \\
p5         &  &   87.73  \\
p7         &  &   85.20  \\
\hline 
p2(+Decoder) &  &  92.08 \\
p7(+Decoder) &  &  88.92 \\
p7(+Decoder) &  &  87.32 \\
\bottomrule
\end{tabular}%
}
\vspace{-2mm}
\end{table}

\subsection{Performances over Different Bag Sizes}
\noindent From table \ref{tab:d}, we have different bag sizes generated with a variation of the parameters lambda and number of training bags to create new copies of dataset. Table \ref{tab:res} shows that as we increase the bag size the performance remains fairly increasing upto a bag size of 8 and then there is a gradual decrease in performance observed with large drop seen with the bag size of 32. Such variation can be attributed to the fact that with the larger bag size the nature of the distribution of the bags to the instance level information per class is artificial and there is a consistent decrease in the performance as we increase bag size from 16 to 32 and beyond. 

\subsection{Effect of Regularizer with Sparse Bags}
In order to analyze the model performance on bags with different sparsity, we further create bags with higher expected counts for each class to generate bags with higher sparsity. The specific setting and data statistics are summarized in Table \ref{tab:d}. Interestingly, we found that the regularized term works really well for sparse bags.  It consistently improve the performance though marginally.

\begin{table}[h]
\centering
\caption{Effect of the Regularizer on Sparse Bags (Dataset ID=p7)}
\label{tab:sparse}
\resizebox{0.45\columnwidth}{!}{%
\begin{tabular}{@{}lll@{}}
\toprule
$\beta$ & Bag Prec. & Inc. Prec. \\ \midrule
0.0                  &    63.70   &    84.67  \\
0.01                 &    64.37  &    85.20  \\
0.1                  &    65.09  &    85.52  \\
0.5                  &    65.07  &    85.87  \\ 
\bottomrule
\end{tabular}%
}
\vspace{-2mm}
\end{table}

\begin{table}[h]
\centering
\caption{Fully supervised upper-bound results with classifier configuration}
\label{tab:fully_supervised}
\resizebox{0.8\columnwidth}{!}{%
\begin{tabular}{@{}llll@{}}
\toprule
Classifier & Loss & Bag Prec. & Inc. Prec. \\ \midrule
ResNet18  & Poisson &    94.26       &   94.40  \\
ResNet34  & Poisson &    94.67       &    94.68  \\
ResNet50  & Poisson &    95.20       &     96.41   \\
MobileNetV2  & Poisson       &    90.03      &     89.07  \\
\bottomrule
\end{tabular}%
}
\vspace{-2mm}
\end{table}

\subsection{Effectiveness of Different Regression Losses}
\label{sec:loss}

 KL loss is mostly widely used loss function for comparing the estimated class distribution and the true class distribution as proposed when learning from proportions. Thus, we use KL-loss as well L1 loss, which is usually used for multi instance regression as alternative choices. 

Table \ref{tab:loss} shows that poisson loss function performs better than other loss functions for bag size equal to 8 and 16, in terms of bag level prediction as well as instance-level prediction. This is as expected as poisson distribution is naturally defined for non-negative and integer values like counting. Another explanation is that the data is generated from poisson distribution but according to Table \ref{tab:sparse}, even though the dataset is generated from non-poisson distribution, the poisson loss function still achieve comparable values. This implies that poisson distribution is robust.

\begin{table}[h]
\centering
\caption{Effect of the Choices of $L_{reg}$ on two standard dataset setting with $N_B\in\{8, 16\}$}
\label{tab:loss}
\resizebox{0.7\columnwidth}{!}{%
\begin{tabular}{@{}ll|l|l@{}}
\toprule
Dataset ID & $L_{reg}$ & Bag Prec. & Inc. Prec. \\ \midrule
p5         & KL        &    90.26       &   90.40  \\
p5         & L1        &    92.67       &    90.68  \\
p5         & Poisson   &    93.20      &     91.21   \\
p3         & KL        &    87.03      &     87.07  \\
p3         & L1        &    92.01       &     85.97  \\
p3         & Poisson   &    92.92       &     88.08  \\ \bottomrule
\end{tabular}%
}
\vspace{-2mm}
\end{table}

\subsection{Comparison with Different Distribution}
Apart from poisson distribution, we also train our models with the exponential and uniform distribution. We observe that the results with the poisson distribution is better as compared the exponential distribution which is in accordance with our expectation as the count information in case of the exponential distribution will be sparse than case of the poisson distribution. The analysis can be reviewed in the Table \ref{tab:res} for case of the dataset id - e0 and p2, there is a stark difference with the same bag size and sparsity distribution but difference in the distribution of classes results in performance drop from 93.2\%  to 90.69\% precision.

\subsection{Variation with the Number of Training Samples}
Additionally, we investigate the degree to which the size of the dataset relates to the model performance. We progressively sample different number of bags and evaluate them on the the test set. More specifically, we compare the results using different scales of training samples. Specific dataset settings are illustrated in Table \ref{tab:loss}.

\begin{table}[h]
\centering
\caption{Scale Testing Results on two standard dataset setting with $N_B\in\{8, 16\}$\}.}
\label{tab:loss}
\resizebox{0.65\columnwidth}{!}{%
\begin{tabular}{@{}llll@{}}
\toprule
Dataset ID & Bag Size & Bag Prec. & Inc. Prec. \\ \midrule
p2         & 8 & 93.20  & 91.21      \\
p5         & 8 & 90.54 & 87.74  \\
p6         & 8 & 77.39 & 68.91      \\ \midrule
\hline
p3         & 16 & 92.93 & 88.08      \\
p8         & 16 & 89.92 & 81.92         \\
p9         & 16 & 84.97 & 69.93          \\ \bottomrule
\end{tabular}%
}
\vspace{-2mm}
\end{table}

\subsection{Variation with the Batch Sizes and the Different Architectures}
According to our experiment, different batch size has different performance. We conduct experiments on variants batch sizes. For all experiments, we train the model with batch size equal to 32, 64, 96, 128 and 256 as seen in Table \ref{tab:backbone} and bag size equal to 8 and 16 as the standard setting. We found that the results with the batch size doesn't provide large variation in the output of the optimization for the bag level training. However, as we increase the batch size there is a drop in performance observed on the instance level prediction.


\begin{table}[h]
\centering
\caption{Ablation Study of Using different backbones and different batch sizes on dataset p2 $B_N=8$.}
\label{tab:backbone}
\resizebox{0.8\columnwidth}{!}{%
\begin{tabular}{lll}
\hline
Backbone          & Bag Prec. & Inc. Prec. \\ \hline
Resnet18 (bs=32)  & 94.036    & 92.096     \\
Resnet18 (bs=64)  & 93.202     & 91.214      \\
Resnet18 (bs=96)  & 93.630    & 91.944     \\
Resnet18 (bs=128) & 93.200     & 91.230       \\
Resnet18 (bs=192) & 92.661    & 90.649     \\
Resnet18 (bs=256) & 92.541    & 90.183     \\
Resnet34 (bs=128)         & 93.763    & 91.763     \\
Resnet50 (bs=128)          & 93.512    & 91.893     \\
MobileNetV2 (bs=128)      &  91.474        &   89.002           \\ \hline
\end{tabular}%
}
\vspace{-2mm}
\end{table}

\section{Conclusion}

In this paper, we propose a novel machine learning problem, namely, learning from counts. We propose a two-stage framework to do instance-level prediction given only a counting vector for a bag is available. We generated dataset from CIFAR10 for experimentation. We achieve comparable results with the fully-supervised setting and much better results than the weakly supervised setting. Additionally, we introduced a L1 regularization term that make our model robust to sparse bags and achieve marginally prediction improvement on sparse bags. We believe semi-weak labels to provide better insights in real world tasks where counting information can be easily obtained and plan to extend our work on other kinds of data.

{\small
\bibliographystyle{ieee_fullname}
\bibliography{egpaper_final}

\begin{thebibliography}{10}\itemsep=-1pt

\bibitem{andrews2003support}
Stuart Andrews, Ioannis Tsochantaridis, and Thomas Hofmann.
\newblock Support vector machines for multiple-instance learning.
\newblock In {\em Advances in neural information processing systems}, pages
  577--584, 2003.

\bibitem{ardehaly2017co}
Ehsan~Mohammady Ardehaly and Aron Culotta.
\newblock Co-training for demographic classification using deep learning from
  label proportions.
\newblock In {\em 2017 IEEE International Conference on Data Mining Workshops
  (ICDMW)}, pages 1017--1024. IEEE, 2017.

\bibitem{babenko2008simultaneous}
Boris Babenko, Piotr Doll{\'a}r, Zhuowen Tu, and Serge Belongie.
\newblock Simultaneous learning and alignment: Multi-instance and multi-pose
  learning.
\newblock 2008.

\bibitem{Baldassarre2020}
Federico Baldassarre, Kevin Smith, Josephine Sullivan, and Hossein Azizpour.
\newblock {Explanation-Based Weakly-Supervised Learning of Visual Relations
  with Graph Networks}.
\newblock In {\em Lecture Notes in Computer Science (including subseries
  Lecture Notes in Artificial Intelligence and Lecture Notes in
  Bioinformatics)}, 2020.

\bibitem{bortsova2018deep}
Gerda Bortsova, Florian Dubost, Silas {\O}rting, Ioannis Katramados, Laurens
  Hogeweg, Laura Thomsen, Mathilde Wille, and Marleen de Bruijne.
\newblock Deep learning from label proportions for emphysema quantification.
\newblock In {\em International Conference on Medical Image Computing and
  Computer-Assisted Intervention}, pages 768--776. Springer, 2018.

\bibitem{carbonneau2018multiple}
Marc-Andr{\'e} Carbonneau, Veronika Cheplygina, Eric Granger, and Ghyslain
  Gagnon.
\newblock Multiple instance learning: A survey of problem characteristics and
  applications.
\newblock {\em Pattern Recognition}, 77:329--353, 2018.

\bibitem{Cheplygina2014}
Veronika Cheplygina, David M~J Tax, Marco Loog, and M~L Oct.
\newblock {On Classification with Bags , Groups and Sets ∗}.
\newblock pages 1--18, 2014.

\bibitem{clements1999subitizing}
Douglas~H Clements.
\newblock Subitizing: What is it? why teach it?
\newblock {\em Teaching children mathematics}, 5(7):400--405, 1999.

\bibitem{Crouse2016OnI2}
D. Crouse.
\newblock On implementing 2d rectangular assignment algorithms.
\newblock {\em IEEE Transactions on Aerospace and Electronic Systems},
  52:1679--1696, 2016.

\bibitem{Datta2019}
Samyak Datta, Karan Sikka, Anirban Roy, Karuna Ahuja, Devi Parikh, and Ajay
  Divakaran.
\newblock {Align2Gound: Weakly supervised phrase grounding guided by
  image-caption alignment}, 2019.

\bibitem{dietterich1997solving}
Thomas~G Dietterich, Richard~H Lathrop, and Tom{\'a}s Lozano-P{\'e}rez.
\newblock Solving the multiple instance problem with axis-parallel rectangles.
\newblock {\em Artificial intelligence}, 89(1-2):31--71, 1997.

\bibitem{doran2014theoretical}
Gary Doran and Soumya Ray.
\newblock A theoretical and empirical analysis of support vector machine
  methods for multiple-instance classification.
\newblock {\em Machine learning}, 97(1-2):79--102, 2014.

\bibitem{dulac2019deep}
Gabriel Dulac-Arnold, Neil Zeghidour, Marco Cuturi, Lucas Beyer, and
  Jean-Philippe Vert.
\newblock Deep multi-class learning from label proportions.
\newblock {\em arXiv preprint arXiv:1905.12909}, 2019.

\bibitem{fan2014learning}
Kai Fan, Hongyi Zhang, Songbai Yan, Liwei Wang, Wensheng Zhang, and Jufu Feng.
\newblock Learning a generative classifier from label proportions.
\newblock {\em Neurocomputing}, 139:47--55, 2014.

\bibitem{foulds2010review}
James~Richard Foulds and Eibe Frank.
\newblock A review of multi-instance learning assumptions.
\newblock 2010.

\bibitem{gao2018c}
Mingfei Gao, Ang Li, Ruichi Yu, Vlad~I Morariu, and Larry~S Davis.
\newblock C-wsl: Count-guided weakly supervised localization.
\newblock In {\em Proceedings of the European Conference on Computer Vision
  (ECCV)}, pages 152--168, 2018.

\bibitem{Gemmeke2017}
Jort~F. Gemmeke, Daniel~P.W. Ellis, Dylan Freedman, Aren Jansen, Wade Lawrence,
  R.~Channing Moore, Manoj Plakal, and Marvin Ritter.
\newblock {Audio Set: An ontology and human-labeled dataset for audio events}.
\newblock In {\em ICASSP, IEEE International Conference on Acoustics, Speech
  and Signal Processing - Proceedings}, 2017.

\bibitem{Hajimirsadeghi2015}
Hossein Hajimirsadeghi, Wang Yan, Arash Vahdat, and Greg Mori.
\newblock {Visual recognition by counting instances: A multi-instance
  cardinality potential kernel}.
\newblock In {\em Proceedings of the IEEE Computer Society Conference on
  Computer Vision and Pattern Recognition}, 2015.

\bibitem{Hernandez-Gonzalez2018}
Jer{\'{o}}nimo Hern{\'{a}}ndez-Gonz{\'{a}}lez, I{\~{n}}aki Inza, Lorena
  Crisol-Ort{\'{i}}z, Mar{\'{i}}a~A. Guembe, Mar{\'{i}}a~J. I{\~{n}}arra, and
  Jose~A. Lozano.
\newblock {Fitting the data from embryo implantation prediction: Learning from
  label proportions}.
\newblock {\em Statistical Methods in Medical Research}, 2018.

\bibitem{7552989}
A. {Kumar} and B. {Raj}.
\newblock Weakly supervised scalable audio content analysis.
\newblock In {\em 2016 IEEE International Conference on Multimedia and Expo
  (ICME)}, pages 1--6, 2016.

\bibitem{letkowski2012applications}
Jerzy Letkowski.
\newblock Applications of the poisson probability distribution.
\newblock In {\em Proc. Acad. Business Res. Inst. Conf}, pages 1--11, 2012.

\bibitem{Liu2019}
Xuejing Liu, Liang Li, Shuhui Wang, Zheng~Jun Zha, Dechao Meng, and Qingming
  Huang.
\newblock {Adaptive reconstruction network for weakly supervised referring
  expression grounding}.
\newblock In {\em Proceedings of the IEEE International Conference on Computer
  Vision}, 2019.

\bibitem{maron1997framework}
Oded Maron and Tom{\'a}s Lozano-P{\'e}rez.
\newblock A framework for multiple-instance learning.
\newblock {\em Advances in neural information processing systems}, 10:570--576,
  1997.

\bibitem{musicant2007supervised}
David~R Musicant, Janara~M Christensen, and Jamie~F Olson.
\newblock Supervised learning by training on aggregate outputs.
\newblock In {\em Seventh IEEE International Conference on Data Mining (ICDM
  2007)}, pages 252--261. IEEE, 2007.

\bibitem{narayan20193c}
Sanath Narayan, Hisham Cholakkal, Fahad~Shahbaz Khan, and Ling Shao.
\newblock 3c-net: Category count and center loss for weakly-supervised action
  localization.
\newblock In {\em Proceedings of the IEEE International Conference on Computer
  Vision}, pages 8679--8687, 2019.

\bibitem{Noroozi2017}
Mehdi Noroozi, Hamed Pirsiavash, and Paolo Favaro.
\newblock {Representation Learning by Learning to Count}.
\newblock In {\em Proceedings of the IEEE International Conference on Computer
  Vision}, 2017.

\bibitem{pappas2014explaining}
Nikolaos Pappas and Andrei Popescu-Belis.
\newblock Explaining the stars: Weighted multiple-instance learning for
  aspect-based sentiment analysis.
\newblock In {\em Proceedings of the 2014 Conference on Empirical Methods In
  Natural Language Processing (EMNLP)}, pages 455--466, 2014.

\bibitem{pappas2017explicit}
Nikolaos Pappas and Andrei Popescu-Belis.
\newblock Explicit document modeling through weighted multiple-instance
  learning.
\newblock {\em Journal of Artificial Intelligence Research}, 58:591--626, 2017.

\bibitem{Patrini2014}
Giorgio Patrini, Richard Nock, Paul Rivera, and Tiberio Caetano.
\newblock {(Almost) no label no cry}.
\newblock In {\em Advances in Neural Information Processing Systems}, 2014.

\bibitem{patrini2014almost}
Giorgio Patrini, Richard Nock, Paul Rivera, and Tiberio Caetano.
\newblock (almost) no label no cry.
\newblock In {\em Advances in Neural Information Processing Systems}, pages
  190--198, 2014.

\bibitem{quadrianto2009estimating}
Novi Quadrianto, Alex~J Smola, Tiberio~S Caetano, and Quoc~V Le.
\newblock Estimating labels from label proportions.
\newblock {\em Journal of Machine Learning Research}, 10(10), 2009.

\bibitem{Rahimi2020}
Amir Rahimi, Amirreza Shaban, Thalaiyasingam Ajanthan, Richard Hartley, and
  Byron Boots.
\newblock {Pairwise Similarity Knowledge Transfer for Weakly Supervised Object
  Localization}.
\newblock In {\em Lecture Notes in Computer Science (including subseries
  Lecture Notes in Artificial Intelligence and Lecture Notes in
  Bioinformatics)}, 2020.

\bibitem{scott2019learning}
Clayton Scott and Jianxin Zhang.
\newblock Learning from multiple corrupted sources, with application to
  learning from label proportions.
\newblock {\em arXiv preprint arXiv:1910.04665}, 2019.

\bibitem{shah2018closer}
Ankit Shah, Anurag Kumar, Alexander~G Hauptmann, and Bhiksha Raj.
\newblock A closer look at weak label learning for audio events.
\newblock {\em arXiv preprint arXiv:1804.09288}, 2018.

\bibitem{Shi2020}
Yong Shi, Jiabin Liu, Bo Wang, Zhiquan Qi, and Ying~Jie Tian.
\newblock {Deep learning from label proportions with labeled samples}.
\newblock {\em Neural Networks}, 128:73--81, 2020.

\bibitem{subramanian2016bayesian}
Saravanan Subramanian, Santu Rana, Sunil Gupta, P~Bagavathi Sivakumar,
  C~Shunmuga Velayutham, and Svetha Venkateshc.
\newblock Bayesian nonparametric multiple instance regression.
\newblock In {\em 2016 23rd International Conference on Pattern Recognition
  (ICPR)}, pages 3661--3666. IEEE, 2016.

\bibitem{tao2004svm}
Qingping Tao, Stephen Scott, NV Vinodchandran, and Thomas~Takeo Osugi.
\newblock Svm-based generalized multiple-instance learning via approximate box
  counting.
\newblock In {\em Proceedings of the twenty-first international conference on
  Machine learning}, page 101, 2004.

\bibitem{tao2004extended}
Qingping Tao, Stephen Scott, NV Vinodchandran, Thomas~Takeo Osugi, and Brandon
  Mueller.
\newblock An extended kernel for generalized multiple-instance learning.
\newblock In {\em 16th IEEE International Conference on Tools with Artificial
  Intelligence}, pages 272--277. IEEE, 2004.

\bibitem{Tsai2019}
Kuen~Han Tsai and Hsuan~Tien Lin.
\newblock {Learning from Label Proportions with Consistency Regularization}.
\newblock {\em arXiv}, pages 1--16, 2019.

\bibitem{vanwinckelen2016instance}
Gitte Vanwinckelen, Daan Fierens, Hendrik Blockeel, et~al.
\newblock Instance-level accuracy versus bag-level accuracy in multi-instance
  learning.
\newblock {\em Data mining and knowledge discovery}, 30(2):313--341, 2016.

\bibitem{wagstaff2007salience}
Kiri~L Wagstaff and Terran Lane.
\newblock Salience assignment for multiple-instance regression.
\newblock 2007.

\bibitem{yu2014modeling}
Felix~X Yu, Liangliang Cao, Michele Merler, Noel Codella, Tao Chen, John~R
  Smith, and Shih-Fu Chang.
\newblock Modeling attributes from category-attribute proportions.
\newblock In {\em Proceedings of the 22nd ACM international conference on
  Multimedia}, pages 977--980, 2014.

\bibitem{yu2013propto}
Felix~X Yu, Dong Liu, Sanjiv Kumar, Tony Jebara, and Shih-Fu Chang.
\newblock Svm for learning with label proportions.
\newblock {\em arXiv preprint arXiv:1306.0886}, 2013.

\bibitem{Zhou2018}
Zhi~Hua Zhou.
\newblock {A brief introduction to weakly supervised learning}, 2018.

\bibitem{Zou2019}
Zhengxia Zou, Wenyuan Li, Tianyang Shi, Zhenwei Shi, and Jieping Ye.
\newblock {Generative adversarial training for weakly supervised cloud
  matting}.
\newblock In {\em Proceedings of the IEEE International Conference on Computer
  Vision}, 2019.

\end{thebibliography}
}

\end{document}